\documentclass[12pt]{article}
\usepackage[utf8]{inputenc}
\usepackage{amsmath, amssymb, amsthm, color, hyperref}
\usepackage{algorithm2e}
\usepackage{graphicx}
\newcommand{\omitthis}[1]{}

\newtheorem{example}{Example}

\newtheorem*{remarknonum}{Remark}

\newcommand{\comment}[1]{}

\newcommand{\eps}{\varepsilon}

\newcommand{\eq}{\begin{equation}}
\newcommand{\eeq}{\end{equation}}

\long\def\omitthis#1{}
\newcommand{\ist}{{\rm({\it i}\,{\rm )}}}
\newcommand{\ind}{{\rm({\it ii}\,{\rm )}}}

\newcommand{\smallbullet}{{\scriptstyle\bullet}}

\makeatletter
\DeclareRobustCommand{\nand}{\mathbin{\mathpalette\n@and@or\land}}
\DeclareRobustCommand{\nor}{\mathbin{\mathpalette\n@and@or\lor}}

\newcommand{\n@and@or}[2]{%
  \vphantom{#2}%
  \ooalign{$\m@th#1#2$\cr\hidewidth$\m@th#1\sim$\hidewidth\cr}%
}

\makeatother

\begin{document}
\title
{\bf On ``Indifference'' and Backward Induction\\ in Games with Perfect Information}
\author{Nimrod Megiddo\thanks{IBM Almaden Research Center, San Jose, California}}
\date{June 2023}
\maketitle

\begin{abstract}
Indifference of a player with respect to two distinct outcomes of a game cannot be handled by small perturbations, because the actual choice may have
significant impact on other players, and cause them to act in a way that has significant impact of the indifferent player. It is argued that ties among  rational
choices can be resolved by refinements of the concept of rationality based on the utilities of other players. One such refinement is the concept of Tit-for-Tat.
\end{abstract}
\section{\hskip -18pt. Introduction}
A common criticism of game theory is that it assumes rationality, and that assumption entails selfishness.  
Thus, if Player $i$ is rational, and has to choose between outcomes $A$ and $B$ with respective utility
payoffs $u_i(A) > u_i(B)$, 
then his choice is must be $A$,
regardless of how small $u_j(A)$, the utility payoff of another player, $j$, may be relative to $u_j(B)$.  
Of course, this criticism can be rebutted by the argument that the utility function $u_i(\smallbullet)$ must take into account {\em all} of the aspects of the outcome and its impact on each of the other players.   
For example, a compassionate player $i$ may prefer to receive a smaller {\em material} payoff if the outcome is much better for another player. 
This preference, however, must be reflected in the utility function $u(\smallbullet)$, so in such a case the outcome with the smaller material payoff to Player $i$ should have a larger utility value to Player $i$.
With this understanding, we assume here as usual, that if Player $i$ has to choose between two different outcomes $A$ and $B$ with utility payoffs $u_i(A) > u_i(B)$, then his choice is must be $A$.

The theorems that state that ``common knowledge of rationality implies backward induction'' 
\cite{aumann1995backward, 
battigalli1999recent,
samet1996hypothetical,
samet2013common,
arieli2015logic,
hillas2020dominance} 
rely on the assumption of {\em no indifference}, which is often called ``genericity'' or ``general position,'' 
meaning  that the preference order over outcomes is a strict total order.
Under this assumption, for every two distinct outcomes $A$ and $B$, and for every player $i$,  
$u_i(A) \not= u_i(B)$. 
The assumption of a strict total order is quite strong. 
An outcome of a game defines the consequences for each individual player.
Naturally, a player may have a strict preference order over consequences for himself but may be indifferent between different consequences for other players.

If the information perfect and no player is indifferent between any two outcomes, then there exists a unique outcome that can be calculated by backward induction.

Here we consider the possibility of indifference, i.e., 
$u_i(A) = u_i(B)$ for some player $i$ for some outcomes $A$ and $B$.
Incidentally, if for two outcomes $A$ and $B$,  $u_j(A)=u_j(B)$ for every player $j$, then $A$ and $B$ may be considered the same outcome, hence the set of distinct outcomes can be much smaller than the set of all leaves of the game tree.
An equality $u_i(A) = u_i(B)$ may seem a ``degenerate'' case.
In other contexts degeneracy can be handled by small perturbations.
Of course, for every utility function $u_i(\smallbullet)$, 
and for every $\eps >0$, there exists a utility function 
$\tilde u_i(\smallbullet)$
such that for every outcome $A$, $| \tilde u_i(A) - u_i(A) | < \eps$, 
and for every  two distinct outcomes $A$ and $B$,
$\tilde u_i(A)\not=\tilde u_i (B)$. 
However, if Player $i$ plays according to $\tilde u_i$, 
then the impact on other players may be significant, 
as we see in examples below.
Therefore, even if Player $i$ may not be too concerned about his choice of action (as reflected in the difference between $u_i$ and $\tilde u_i$), 
{\em other players may have to rationalize carefully about what Player $i$ might do}.
Because Player $i$ may, in turn, be severely impacted by other players' actions,  Player $i$ should also try to analyze carefully about how the other players
rationalize about his own choices.
Therefore, {\em small tie-breaking perturbations do not lead to a reasonable decision-making procedure}.

We attempt to provide here a different approach to indifference.

\section{\hskip -18pt. Acting under indifference}
If Player $i$ must choose one of $k$ alternatives $A_1,\ldots, A_k$, where
\[ u_i(A_1) = u_i(A_2) > u_i(A_3) \ge u_i(A_4) \ge \cdots \ge u_i(A_k) ~,\]
then a rational choice is ether $A_1$ or $A_2$.
Rationality, however, does not determine anything more than that. 
Therefore, the combination of rationality and perfect information does not suffice for determining the decisions players make.

In previous work \cite{megiddo2010towards, megiddo1} we argued that preferences (hence also utility functions) may change during a play of a game, depending on the actions of other players. 
Thus, the welfare of one player may be factored into the utility function of another player, positively or negatively, depending on the past actions of the former.

Here, we consider a different aspect of making decisions in a game.
We assume the utility function does not change during the play, 
but the choice of one from all available rational actions may depend on the past actions of other players.
This issue can be interpreted as a refinement of the set of outcomes.
In the formal description of the game, each leaf of the game tree has an associated outcome, 
and the utility function is defined over the set of outcomes.
Two outcomes that have the same utility for one player, 
may be associated with endpoints that differ in terms of the moves that lead to these endpoints. 
Thus, a player may have a secondary preference relation that resolves
ties that exist in the primary preference relation that is behind the given
utility payoffs. 
That secondary preference relation may guide the choices when the primary relation does not.  
It is interesting to note that if the description of the game should be the only basis for reasoning about choices, then the secondary preference relation of a player should be derived only from the utility values of the {\em other} players.

\begin{example} \rm
Consider a two-person game of the type depicted in Figure \ref{twop}, which we will call {\em Indifference}.
Obviously, Player 2 is indifferent between the two choices at each of his decision nodes.
\begin{figure}  [ht]
\begin{center}
\includegraphics[width=5in]{"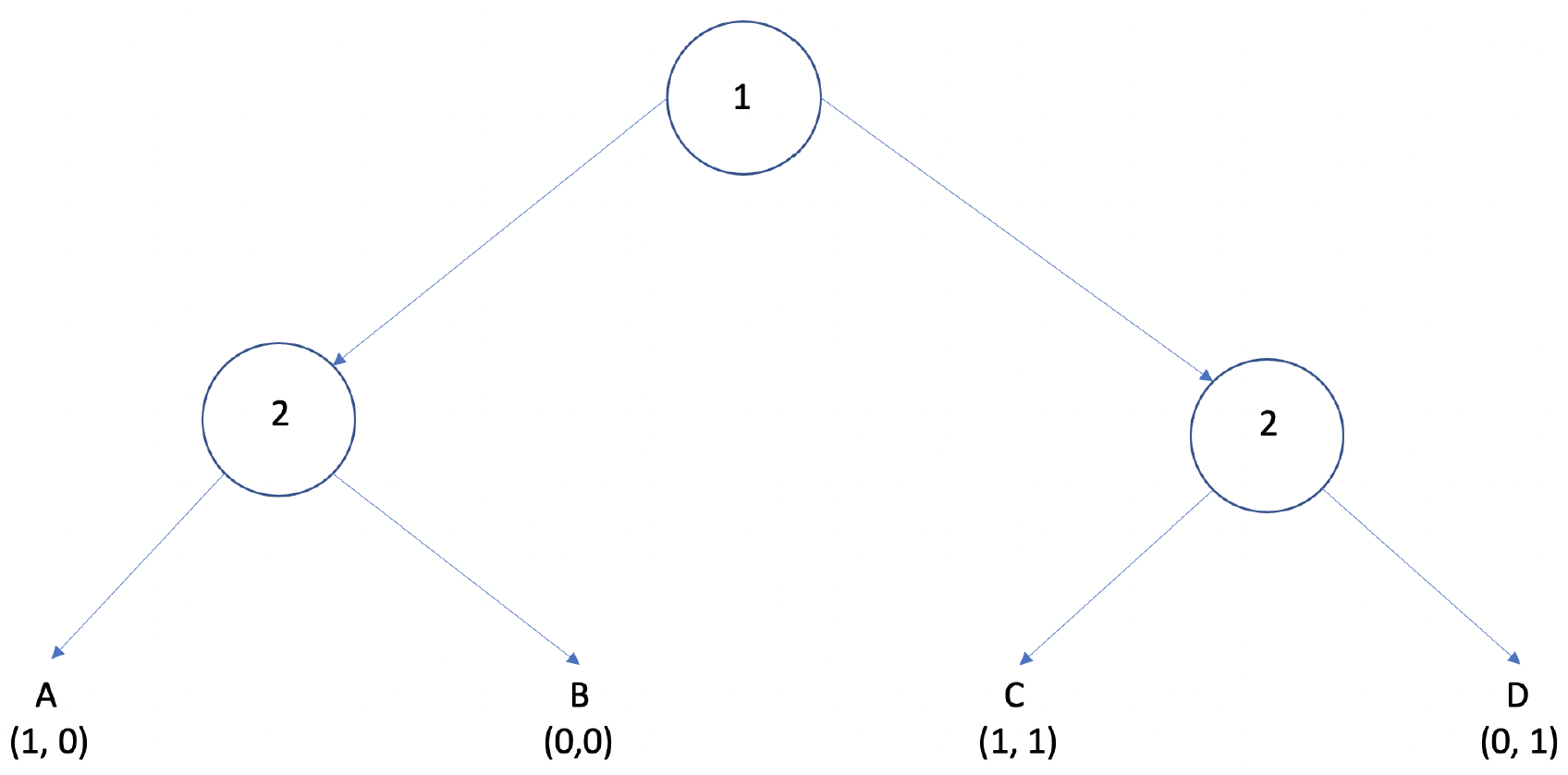"} 
\caption{Indifference game} \label{twop}
\end{center}
\end{figure}
Furthermore, Player 1 has to choose between the two decision nodes of Player 2, which have the following properties:
\ist\ 
in each of the decision nodes of Player 2, if Player 2 chooses Left, then Player 1 gets 1, and if he chooses Right, then Player 1 gets 0, and
\ind\ 
in each of them Player 2 is indifferent between choosing Left or Right.
Therefore, it seems that both players are indifferent. 
There is, however, some subtlety here as we discuss below.
\end{example}

\subsection{Characterizing choices as friendly or unfriendly}
Given that  in the Indifference game (Figure \ref{twop})  Player 2 is indifferent between 
the outcomes $A$ and $B$, we say that choosing $A$ is 
{\em friendly}, while choosing $B$ is {\em unfriendly}. 
Similarly, given that Player 2 is indifferent between 
the outcomes $C$ and $D$, we say that choosing $C$ is 
friendly, while choosing $D$ is unfriendly. 
\begin{remarknonum} \rm
The characterization of an action as friendly can be generalized to ultimate-decision nodes with arbitrary numbers of choices, where the player has at least two best  choices. It can further be generalized to any number of players.
Recall that we assume the players are rational, i.e., each player always chooses action that is best available for himself or herself.
However, a player may still have to choose one from his best available actions.
In general, we say that a rational choice of Player $j$ is {\em friendly to Player $i$} if it is best for Player $i$ among the choices that are best for Player $j$;  otherwise, it is {\em unfriendly to Player $i$}. Alternately, we may quantify the extent to which Player $j$'s choice is friendly to Player $i$.  
The ``quantified'' friendliness may be useful for Player $i$ if the latter wishes to compare two other players, $j$ and $k$, whose choice is more friendly to Player $i$.
The comparison may guide Player $i$ as to in whose favor he should resolve ties in his own choices.
\end{remarknonum}

In the Indifference game, Player $1$ evaluates the two nodes, between which he has to choose, under different scenarios of Player $2$'s choices.  
If Player $2$ plays friendly at the AB node, 
then Player $1$ receives there (i.e., at A) a payoff of $1$; otherwise, Player $1$ receives there (i.e., at B) a payoff of $0$. 
If Player $2$ plays friendly at the CD node, 
then Player $1$ receives there (i.e., at C) a payoff of $1$; otherwise, Player $1$ receives there (i.e., at D) a payoff of $0$. 
Thus, it seems that Player $1$ is indifferent between the two choices, 
hence his decision could be based on the utility of Player $2$.
Note that if Player $1$ chooses the CD node, 
then Player $2$ receives a payoff of $1$, 
and if Player $1$ chooses the AB node, 
then Player $2$ receives a payoff of $0$. 
Therefore, choosing the CD node can be deemed friendly, whereas choosing the AB node can be deemed unfriendly.

Table \ref{one-step} is the normal-form representation of the game, 
F is the friendly strategy of Player 1, 
U is the unfriendly strategy of Player 1.
Player 2 has a strategy FU  (which may be called ``Tit-for-Tat'') according to which Player $2$ plays friendly if and only if Player $1$ played friendly. 
UF is Player 2' s strategy of playing friendly if and only if
Player 1 does {\em not} play friendly, 
FF is Player 2' s strategy of always playing friendly, 
and UU is Player 2' s strategy of always playing unfriendly.
\begin{table}   \label{one-step}
\begin{center}
\begin{tabular}{|c ||c c | c c | c c | c c |}
\hline
   & FU & (BC) & UF& (AD)& FF& (AC) &UU & (BD) \\
\hline \hline
F  &   & 1  & & 1    &   & 1 &   & 1 \\    
(CD)   & 1 &    & 0 &    & 1 &   & 0 &   \\   
\hline
U  &   & 0  & & 0   &  & 0 & & 0\\    
(AB)   & 0 &    & 1 &   &1 &   & 0 & \\   
\hline
\end{tabular}
\caption{One-step perfect information} \label{one-step}
\end{center}
\end{table}
Obviously, the columns of player 2' payoff matrix are identical, which may mislead to believe that all the choices Player 2 has are equivalent.
Note that Player 1's best response to Player 2's Tit-for-Tat is to play friendly, which gives rise to the outcome of $(1,1)$ in equilibrium.

The pure Nash-equilibria in this game are
\begin{center}
(F, FU), (F, FF), (F, UU), (U, UF), (U, FF), (U, UU).
\end{center}

Note that FU is better for Player $2$ than FF, 
because the former guarantees a payoff of $1$ to Player 2 if Player $1$ is rational, 
whereas the latter results in a payoff of $1$ to Player $2$ only if Player $1$ is friendly.

\section{\hskip -18pt. A multi-stage version}
A multi-stage extension of the Indifference game can be defined with a simple interpretation as follows. 
We can define either a repeated game of imperfect information (because players move simultaneously), 
or a game of perfect information in which players move alternately.

In the (finite or infinite) repeated-game version, the stage game is Simultaneous Indifference (shown in Table \ref{rep:game}).
\begin{table}
\begin{center}
\begin{tabular}{|c||ccc|ccc|} 
\hline
 &  & F & &   & U & \\
\hline
\hline
  &   && 1 &   & &1 \\
F &   &&   &   & &  \\
 & 1 &&   & 0 & &   \\
 \hline
  &   && 0 &   && 0 \\
U &&&&&&\\
  & 1 &&   & 0 &&   \\
 \hline
\end{tabular}
\end{center}
\caption{The Simultaneous Indifference game} \label{rep:game}
\end{table}
  
In this stage game, each player essentially decides whether the other player will receive a unit of utility.
Because none of the players can affect its own utility, each of the pairs  (F, F), (F, U), (U, F) and (U, U) is a Nash-equilibrium in the stage game.
Of course, the most desirable for both players is (F, F) but indifference of one player is risky for the other player.
The Tit-for-Tat strategy in the repeated game is to start with F and the at each stage play exactly what the other player played in the preceding stage.
In the repeated game, the Tit-for-Tat strategy of one player incentivizes the other player not to be indifferent.

A closely-related perfect-information game is the one where players decide, alternately, whether the other player will receive a unit of utility.
The ``Tit-for-Tat'' strategy in this game is essentially the same in the simultaneous moves version, i.e., to play the same as the other player in the preceding move or F if this is the very first move in the game.

\section{\hskip -18pt. Extending backward induction}
The essential characteristic of backward induction is that the decision at a node depends only on the subgame rooted at that node.  
Once the optimal decision at a node is determined, the subgame rooted at the node can be collapsed into single node.
Thus, the whole process reduces to a sequence of  ultimate decisions, i.e., at nodes all of whose children are leaves.

As noted above, when a player has more than one best choice, then there may exist other factors (i.e., in addition to rationality) that guide the actual choice of how to break the ties among rational choices.
We wish to incorporate such factors in a backward-induction-like procedure.  
However, the particular values of such factors may depend on the interpretation of actions along the path from the root to the decision node and even on
other parts of the tree. 
Thus, the node may be considered under various scenarios, and later, when the relevant scenario is identified, the decision at that node can be readily available.
 \begin{example} \label{ex:1} \rm
 Consider the decision tree depicted in Figure \ref{factors}.
 \begin{figure}  [ht]
\begin{center}
\includegraphics[width=5in]{"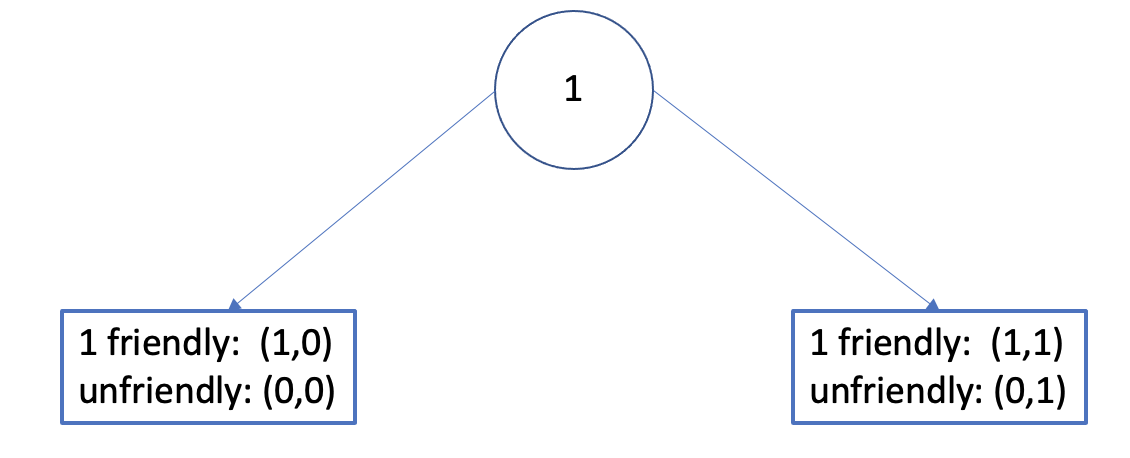"} 
\caption{Tentative values} \label{factors}
\end{center}
\end{figure}
This tree represents the outcomes when Player 2 plays friendly if and only if Player 1 does.
Recall that this strategy of Player $2$ is only one of several rational strategies, hence it is not necessarily  the one selected by the backward induction process.
Note that 
\ist\ Player 1 receives $1$ if he plays friendly and $0$ if not, and
\ind\ if Player 1 chooses Left, then Player 2 receives $0$, and if he chooses Right, then Player 2 receives $1$.
Therefore, Player 1's choosing Right is friendly and leads to the outcome of $(1,1)$, 
whereas  Player 1's  choosing Left is unfriendly and leads to the outcome of $(0,0)$.
Thus, in this circumstance the only rational choice of Player 1 is to play friendly.
\end{example}

If there is only one player, then a tie can be resolved arbitrarily or randomly. 
The presence of other players complicates the analysis,
because the particular way of resolving ties may impact other players and cause them to act in certain ways that may affect the tie-breaking player.
One possibility for the tie-breaking player is to consider the decision in various scenarios of attitudes towards other players.
More precisely, in the special case of two players, the deciding player may consider two scenarios, namely, the other player
is deemed friendly or unfriendly. 
If the other player is deemed friendly, then the deciding player may resolve a ties in favor of the other player; otherwise, he may
resolve it in an unfriendly way.
The question of whether the other player should be deemed friendly can be addressed later but the idea is that for backward induction
 tentative choice can be selected for each possible scenario.
 \begin{example} \label{ex:2}  \rm
 Consider the game depicted in Figure \ref{threestage}.
 \begin{figure}  [ht]
\begin{center}
\includegraphics[width=5in]{"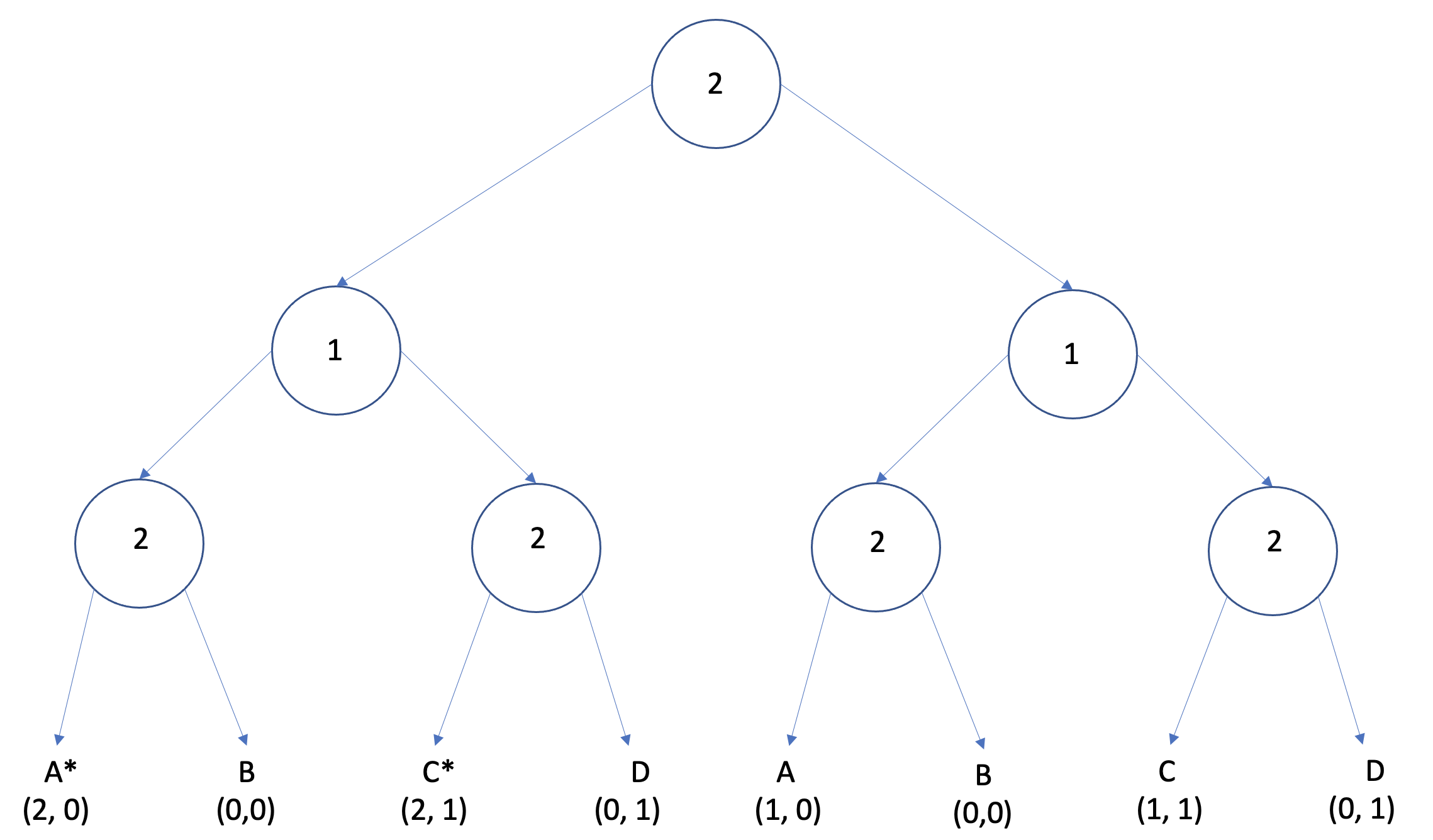"} 
\caption{Three-stage indifference game} \label{threestage}
\end{center}
\end{figure}
Here, Player 2 makes the first move, deciding which of two subgame is subsequently played.
Each of these subgames is essentially the same as the Indifference game of Figure \ref{twop}.
We already saw in Example \ref{ex:1} that if Player 2 plays the Tit-for-tat strategy, the the only rational strategy
of Player 1 is to play friendly.  
Therefore, in the three-stage game, no matter what Player 2 chooses in the first stage,
the only rational choice of Player 1 is to play friendly.  
Note that Player 2 is indifferent in the first stage.
In fact, the collapsed subtrees present Player the choice between $(2,1)$ from the left-hand side or $(1,1)$ from the right-hand side.
Player 2 may play friendly by choosing Left or unfriendly by choosing Right.
The tie can be resolved arbitrarily, but i this is a subtree of a bigger tree, then the choice may depend on how Player 1 has played
before this subtree was reached.

If Player 1 is rational, the choice of Player 2 in the first stage does not matter for Player 1.
Of course, Player 1 may threaten to play unfriendly if Player 2 chooses Right, but this threat is not credible if Player 1 is rational.

\end{example}

\bibliographystyle{abbrv}

\end{document}